\documentclass[11pt]{article}
\PassOptionsToPackage{table}{xcolor}
\usepackage[preprint]{acl}

\usepackage{times}
\usepackage{latexsym}

\usepackage[T1]{fontenc}

\usepackage[utf8]{inputenc}

\usepackage{microtype}

\usepackage{inconsolata}

\usepackage{graphicx}
\usepackage{amsmath,amssymb}   
\usepackage{algorithm}
\usepackage{algpseudocode}
\usepackage{booktabs,multirow}
\usepackage[table]{xcolor}
\usepackage{arydshln}
\usepackage{enumitem}
\usepackage{amsthm}
\usepackage{url}
\theoremstyle{definition}
\newtheorem{definition}{Definition}
\renewcommand{\arraystretch}{1.3}  %

\title{TimesJev: Thinking Fast on Time Series with System 1 Sufficiency}

\author{
  \textbf{Chenghao Liu\textsuperscript{1}},
  \textbf{Wei Fan\textsuperscript{1}},
  \textbf{Jindong Han\textsuperscript{2}},
  \textbf{Dongjie Wang\textsuperscript{3}},
  \\
  \textbf{Aniket Mahanti\textsuperscript{1}},
  \textbf{Pengyang Wang\textsuperscript{4}},
  \textbf{Hao Liu\textsuperscript{5}}
  \\
  {\normalfont
    \textsuperscript{1}University of Auckland,
    \quad
    \textsuperscript{2}Shandong University}
  \\
  {\normalfont
    \textsuperscript{3}Northeast Normal University,
    \quad
    \textsuperscript{4}University of Macau}
  \\
  {\normalfont
    \textsuperscript{5}The Hong Kong University of Science and Technology (Guangzhou)}
  \\
  {\small\normalfont\ttfamily
    \begin{tabular}{@{}c@{}}
      \{chenghao.liu,wei.fan,a.mahanti\}@auckland.ac.nz,
      jindong.han@sdu.edu.cn, \\
      wangdongjie100@nenu.edu.cn, pywang@um.edu.mo, liuh@hkust-gz.edu.cn
    \end{tabular}
  }
}

\begin{document}
\maketitle
\begin{abstract}
Large language models (LLMs) have recently shown strong potential for time series reasoning, while models that reason over time series are usually with long chains of thought. However, this process often consumes thousands of tokens, directly leading to substantial latency during inference. In contrast, System 1 models such as Jev make decisions instantly and can naturally be an alternative, but raw time series are not compatible with the embedding space of these models, thus hindering the reasoning performance, especially when input series contain thousands of numeric points. In this paper, we make a very early attempt to propose a time series System 1 model, namely \textit{TimesJev}. Specifically, we first formalize System 1 sufficiency in time series reasoning, which motivates a compressed representation for time series with key frequency patterns preserved. 
To construct this representation, we derive a dimension reduction objective that combines (i) a sufficiency divergence that measures the core information lost during compression, and (ii) a realizability divergence that quantifies the incompatibility of the System 1 model to the optimal prediction given this representation.
We optimize this objective by learning time series compression with a gating mechanism for sufficiency, and then aligning the selected representation with the embedding space for realizability.
Finally, experiments on 13 datasets demonstrate that TimesJev outperforms reasoning models by over 62\% with 62.9\% less inference time than their fine-tuned versions on average. Our code and models are publicly available.\footnote{Code: \url{https://github.com/coenliu/TimesJev}; models: \url{https://huggingface.co/Coenliu62/TimesJev}}
\end{abstract}

\section{Introduction}
Time series reasoning aims to analyze numerical data to address complex questions and support tasks such as forecasting~\cite{qiu2026dbloss,fan2022depts}, classification~\cite{liu2025multiscale}, and analytical question answering~\cite{zhou2025time}, with growing importance in applications spanning healthcare~\cite{noroozizadeh2026forecasting}, energy~\cite{ye2025llm}, and climate~\cite{zhou2025eternagram}.
Building on this progress, recent work has explored explicit reasoning to improve the performance of large language models (LLMs) in time series tasks, through reinforcement learning with long chains of thought~\cite{guan2026timeomni}, supervised fine-tuning on stepwise reasoning traces~\cite{regmi2026timethink}, or agent frameworks that alternate between reasoning steps and tool calls~\cite{liu2025ts}.

Although these explicit reasoning approaches generally improve time series reasoning performance, they require generating thousands of tokens for each question during reasoning, which substantially increases inference latency. This becomes even more severe for long time series data, where LLMs need to generate longer reasoning chains to understand the series patterns~\cite{cheng2026can}.
As shown in Figure~\ref{fig:motivation}, reasoning models incur higher inference latency with longer reasoning in time series reasoning tasks, without a corresponding gain in accuracy.
However, System 1 models, such as Jev~\cite{typesafe2026api}, directly make decisions from the question and the series in a single forward pass to avoid generating expensive reasoning steps~\cite{snell2025scaling}. Such a model returns answers without generating explicit steps, which significantly reduces inference latency.

While System 1 models can efficiently make decisions under limited test-time computation, identifying patterns in noisy time series remains a challenge for them~\cite{merrill2024language}. Under such a limitation, the model has to understand long series that are tokenized into discrete tokens and implicitly infer a pattern, such as a trend or periodicity~\cite{gruver2023large,fan2023dish}. Noisy series further make this recognition unreliable, and small deviations in the recognized patterns can lead to a completely different final answer~\cite{zhou2025can}. Moreover, these challenges are confirmed in Jev's report, which identifies processing numerical input as a weakness~\cite{typesafe2026api}.

To address these challenges, we explore a novel approach to sufficiently compressing time series for System 1 models such as Jev. The core idea of this approach, \textit{TimesJev}, is to learn a sufficient representation of the time series and enable the model to leverage it effectively. Specifically, we first formalize System 1 sufficiency in time series reasoning, which motivates compressing each series into a compact representation that preserves the essential information.
To build such a representation, we introduce a dimension reduction objective that combines a sufficiency divergence, which measures the task-relevant information of the series lost after compression, and a realizability divergence, which measures how far the System 1 model is from the optimal prediction given this representation.
To approach sufficiency with a lower-dimensional representation, we learn a selector with stochastic gates to retain a small set of temporal features without losing the core information. In order to align the representation with the model, we project the compact representation into input embeddings that the model can use for accurate direct decisions.
In summary, our contributions are as follows:
\begin{itemize}[leftmargin=*]

\item We explore a novel direction of \textit{time series reasoning with System 1 sufficiency}. Unlike existing methods that directly process the raw time series, we compress it into a lower-dimensional representation for direct decision making by System 1 models. We introduce a dimension reduction objective that measures both the sufficiency of this representation and its realizability for the model.

\begin{figure}[t]
\centering
\includegraphics[width=0.95\columnwidth]{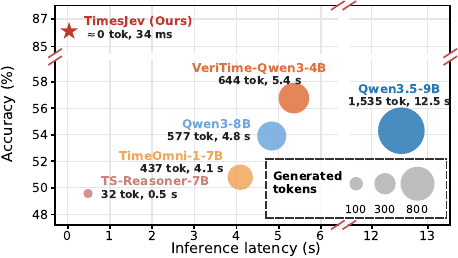}
\caption{Accuracy versus inference latency on time series tasks, with bubble size denoting generated tokens.}
\label{fig:motivation}
\end{figure}

\item We propose \textit{TimesJev}, a time series System 1 model that learns a sufficient representation of the time series. To approach sufficiency with a lower-dimensional representation, we utilize stochastic gates, which retain a subset of temporal representation with the most predictive information. To approach realizability, we further propose a lightweight encoder which projects the compact representation into the input embedding space of the model.

\item We conduct experiments on TSQA~\citep{kong2025time}, covering 13 datasets across 4 tasks, showing that TimesJev outperforms existing time series reasoning methods. TimesJev achieves an average relative improvement of more than 69\% in overall accuracy over reasoning models, while reducing latency by 62\% on average compared with their supervised fine-tuned versions.
\end{itemize}

\section{Related Works}

\paragraph{Time Series Reasoning.}
Time series reasoning aims to identify and interpret temporal patterns in a series to solve tasks such as anomaly detection, classification, and question answering~\cite{kong2025achieving,kong2025time}. Current work follows three main directions to enhance the time series reasoning ability of LLMs~\cite{chang2025survey}. One line trains LLMs with reinforcement learning that rewards correct answers, typically after supervised fine-tuning~\cite{guan2026timeomni,zhou2025time,zhou2026time}. Another line fine-tunes LLMs on step-by-step reasoning traces, such as structured reasoning annotations for anomaly diagnosis~\cite{yang2026time,ye2026surveyanomaly} or traces corrected by a critic model~\cite{su2025chain}. The third line employs LLMs as agents that interleave reasoning steps with time series analysis tools in a ReAct-style loop~\cite{yao2022react,liu2025ts}, and some agents are further trained on expert tool-use trajectories~\cite{wu2026timeart}.

\paragraph{Time Series Encoding.}
To enable LLMs to better process time series data, a line of work maps the series into the input space of the model. The most direct approach serializes the values as text, either as digit strings at a fixed precision~\cite{gruver2023large} or as discrete tokens added to the vocabulary after normalization~\cite{wang2025chattime}. Another line learns patch embeddings and aligns them with the word embedding space, for example, through text prototypes or semantic anchors, while keeping most of the LLM parameters frozen~\cite{zhou2023one,jin2024time,pan2024s,cao2024tempo,liu2024autotimes}. Time-LLM also adds input statistics to the prompt as text~\cite{jin2024time}. For reasoning, recent studies attach a time series encoder to the LLM, ranging from a lightweight network to a pretrained time series foundation model, and feed its outputs as tokens~\cite{yu2025ts}.

\paragraph{System 1 Models.}
Recent studies on LLM reasoning distinguish fast System 1 models, which produce the answer directly without intermediate tokens, from System 2 models, which generate intermediate tokens, such as a chain of thought, before the answer~\cite{yu2024distilling,zhang2025system,li2026jeva,liyubo2026jev}.
Reasoning models of the latter kind tend to overthink, generating redundant tokens even for simple questions~\cite{chen2025not,li2026sponge}. 
Meanwhile, chain-of-thought reasoning brings large gains on math and symbolic tasks, with much smaller gains on other tasks~\cite{sprague2025cot}. 
Even in such tasks, reasoning models can answer more accurately without thinking with the same output length~\cite{ma2025reasoning}.
To leverage the strengths of System 1 models under limited test-time computation, prior work distills the outputs of System 2 methods into models that answer without intermediate steps~\cite{yu2024distilling}, or lets models state the answer first and justification afterwards~\cite{xuan2026post}. Moreover, LLMs can reason internally without writing out every step~\cite{baherwani2026not}, suggesting that System 1 models can still perform well without explicit reasoning chains.

\section{Problem Formulation}
This paper studies the task of \textit{time series reasoning}~\cite{chow2024towards}, which requires answering questions over time series data. 
Formally, given time series observations $\mathcal{X}=(x_1,\ldots,x_n)$, where $x_t\in\mathbb{R}^d$, and a textual context $\mathcal{C}$ containing a question and candidate answers $\mathcal{A}(\mathcal{C})$, time series reasoning aims to predict the correct answer $\mathcal{Y}\in\mathcal{A}(\mathcal{C})$ by identifying relevant temporal patterns and reasoning from them. 
We study this task with a language model $q_0$ that predicts an answer in a single forward pass, without generating explicit intermediate reasoning steps:
\begin{equation}
    \hat{\mathcal{Y}}
    = \operatorname*{arg\,max}_{y\in\mathcal{A}(\mathcal{C})} q_0(y\mid \mathcal{C},\mathcal{X}).
    \label{eq:task}
\end{equation}

\begin{figure*}[t]
\centering
\resizebox{\textwidth}{!}{%
\includegraphics{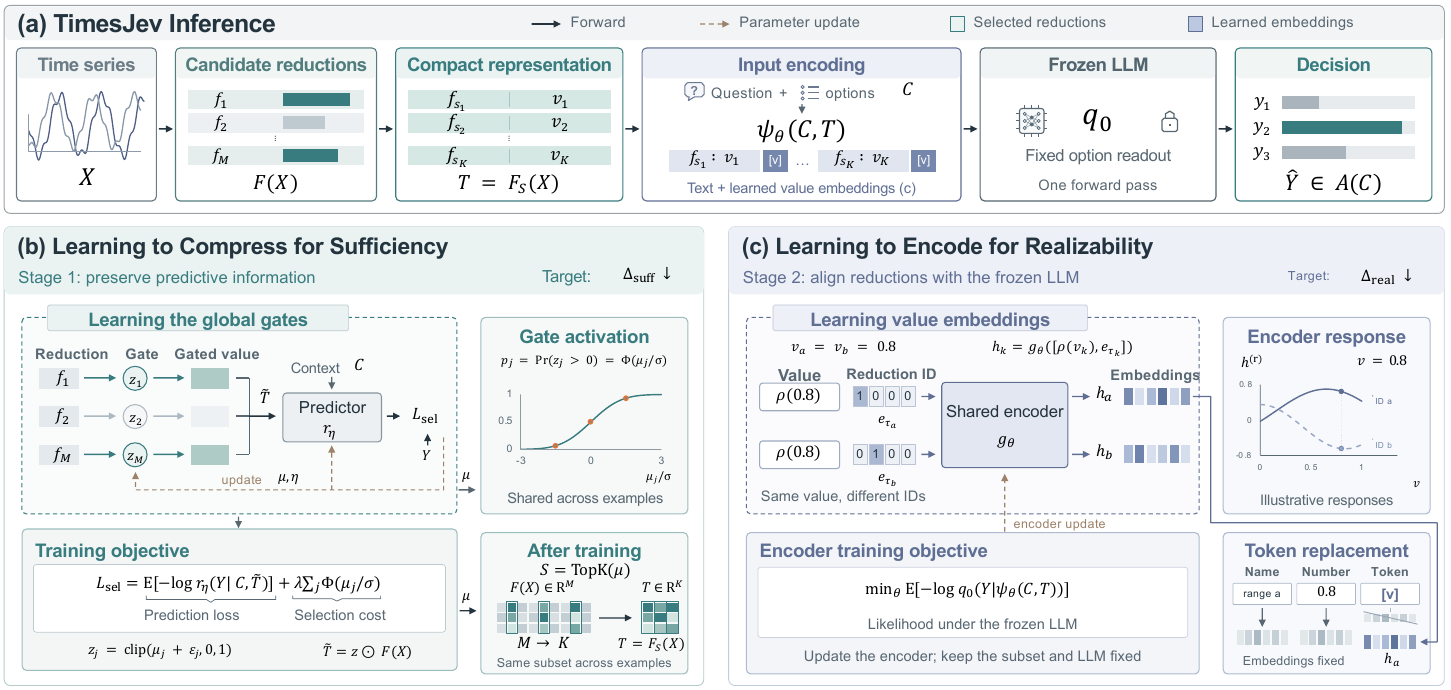}%
}
\caption{Overview of TimesJev. (a) Inference with a System 1 model in one forward pass. (b) Compressing the time series through stochastic gates to reduce the sufficiency divergence. (c) Aligning the compact representation with the model through learned embeddings to reduce the realizability divergence.}
\label{fig:method}
\end{figure*}

\section{TimesJev}
In this section, we present TimesJev, which learns a sufficient representation of the time series for direct decision making by System 1 models, as shown in Figure~\ref{fig:method}. We first formalize System 1 sufficiency of time series representations (Section~\ref{sec:reduction}).
Based on this, we learn to compress each series into a compact representation that approaches sufficiency (Section~\ref{sec:compression}), and to encode this representation for realizability by the System 1 model (Section~\ref{sec:embedding}).

\subsection{System 1 Sufficiency of Time Series Representations}
\label{sec:reduction}
System 1 models such as Jev~\cite{typesafe2026api} directly return decisions in a single forward pass without generating explicit reasoning steps. However, applying this paradigm to time series reasoning requires identifying time series patterns from long and noisy numerical data, which can introduce errors in series pattern recognition~\cite{liu2025towards} and, even worse, lead to incorrect answers under such limited test-time computation. This motivates a compact time series representation that characterizes these patterns and is sufficient to answer the question. Given this, the model no longer needs to infer the patterns from the raw series. Moreover, the model should be able to use this representation effectively for direct decisions. We call a representation that satisfies both requirements System 1 sufficient, and formalize them below.

\begin{definition}[System 1 Sufficiency]
\label{def:system1_sufficiency}
Let $\Psi$ be a family of input encoders, each mapping the context $\mathcal{C}$ and a representation $\mathcal{T}=\phi(\mathcal{X})$ to an input of the System 1 model $q_0$. The representation $\mathcal{T}$ is System 1 sufficient for the answer $\mathcal{Y}$ given $\mathcal{C}$ if there exists $\psi\in\Psi$ such that
\begin{align}
    p(\mathcal{Y}\mid\mathcal{C},\mathcal{T}) &= p(\mathcal{Y}\mid\mathcal{C},\mathcal{X}), \label{eq:conditional_sufficiency}\\
    q_0\big(\mathcal{Y}\mid\psi(\mathcal{C},\mathcal{T})\big) &= p(\mathcal{Y}\mid\mathcal{C},\mathcal{T}), \label{eq:realizability_condition}
\end{align}
where $p$ denotes the underlying data distribution.
\end{definition}
Below, we measure the violation of Eq.~\eqref{eq:conditional_sufficiency} and Eq.~\eqref{eq:realizability_condition} by the KL divergence between the two sides of each, which we call the sufficiency divergence and the realizability divergence, respectively.

\paragraph{Predictive Sufficiency under Compression.}
The requirement to preserve the time series information needed to answer the question can be formalized through sufficient dimension reduction (SDR)~\cite{park2010dimension}. Classical SDR seeks a lower-dimensional linear projection of the observations that preserves the conditional distribution of the prediction target. \citet{lee2013general} generalize this formulation to nonlinear functions, providing a broader class of functions for constructing sufficient representations. In particular, such nonlinear functions can characterize temporal patterns in time series data, including variability, periodicity, and lagged dependence~\cite{cheng2015time}.

Building on this, we formulate a dimension reduction problem to obtain a compact time series representation $\mathcal{T}=\phi(\mathcal{X})$ that satisfies the sufficiency condition in Eq.~\eqref{eq:conditional_sufficiency}, i.e., preserves all information in $\mathcal{X}$ relevant to predicting the answer given $\mathcal{C}$. However, under a finite representation, a compact representation may discard some of this information. To quantify the resulting loss, $H(\mathcal{Y}\mid \mathcal{C},\mathcal{X})$ measures the uncertainty about the answer given the complete observations, whereas $H(\mathcal{Y}\mid \mathcal{C},\mathcal{T})$ measures the uncertainty remaining after replacing $\mathcal{X}$ with $\mathcal{T}$. For a deterministic representation $\mathcal{T}$, their difference defines the sufficiency divergence:
\begin{equation}
\begin{aligned}
    \Delta_{\mathrm{suff}}
    &= H(\mathcal{Y}\mid \mathcal{C},\mathcal{T})-H(\mathcal{Y}\mid \mathcal{C},\mathcal{X})\\
    &= \mathop{\mathbb{E}}\limits_{\mathcal{C},\mathcal{X}}\mathrm{KL}\bigl(p(\cdot\mid\mathcal{C},\mathcal{X})\,\Vert\,p(\cdot\mid\mathcal{C},\mathcal{T})\bigr),
\end{aligned}
\label{eq:compression_loss}
\end{equation}
where $H$ denotes entropy and $\mathrm{KL}$ the Kullback--Leibler divergence; $\Delta_{\mathrm{suff}}$ also equals the conditional mutual information $I(\mathcal{X};\mathcal{Y}\mid\mathcal{C},\mathcal{T})$. The divergence is zero exactly when $\mathcal{T}$ satisfies Eq.~\eqref{eq:conditional_sufficiency}. Minimizing this divergence under the representation budget therefore provides an objective for preserving the core information to answer question.

\paragraph{Realizability by System 1 Models.}
The previous divergence quantifies the predictive information lost in dimension reduction. However, preserving this information does not ensure that the LLM can use it effectively to predict the answer. To connect the representation to the LLM, each input encoder $\psi\in\Psi$ in Definition~\ref{def:system1_sufficiency} maps the question context $\mathcal{C}$ and time series representation $\mathcal{T}$ to a complete model input $s=\psi(\mathcal{C},\mathcal{T})$. The LLM, together with its fixed option readout, produces a prediction distribution $q_0(\cdot\mid s)$ over the candidate answers. For a given representation $\mathcal{T}$, the optimal expected log loss within this family is:
\begin{equation}
    R_\Psi^*(\mathcal{T})
    =
    \inf_{\psi\in\Psi}
    \mathbb{E}\!\left[
        -\log q_0\!\left(\mathcal{Y}\mid\psi(\mathcal{C},\mathcal{T})\right)
    \right].
    \label{eq:constrained_loss}
\end{equation}

Given $(\mathcal{C},\mathcal{T})$, an unrestricted predictor can attain the minimum expected log loss $H(\mathcal{Y}\mid \mathcal{C},\mathcal{T})$. Model and interface constraints introduce a realizability divergence:
\begin{equation}
\begin{aligned}
    \Delta_{\mathrm{real}}
    &= R_\Psi^*(\mathcal{T})-H(\mathcal{Y}\mid \mathcal{C},\mathcal{T})\\
    &= \inf_{\psi\in\Psi}\mathop{\mathbb{E}}\limits_{\mathcal{C},\mathcal{T}}\mathrm{KL}\bigl(p(\cdot\mid\mathcal{C},\mathcal{T})\,\Vert\,q_0(\cdot\mid s)\bigr).
\end{aligned}
\label{eq:realizability_loss}
\end{equation}
This divergence measures the excess prediction loss that remains even with the best input encoder in $\Psi$. It is zero when Eq.~\eqref{eq:realizability_condition} holds, i.e., when the posterior $p(\mathcal{Y}\mid\mathcal{C},\mathcal{T})$ is realizable by the System 1 model, as in the realizable setting of learning theory~\citep{shalev2014understanding}.

Together, sufficiency and realizability both affect predictive performance. Combining this with the sufficiency divergence in Eq.~\eqref{eq:compression_loss} gives:
\begin{equation}
    R_\Psi^*(\mathcal{T})-H(\mathcal{Y}\mid \mathcal{C},\mathcal{X})
    =
    \Delta_{\mathrm{suff}}+\Delta_{\mathrm{real}}.
    \label{eq:risk_decomposition}
\end{equation}
Hence, for a System 1 sufficient $\mathcal{T}$, both divergences are zero, and the System 1 model attains $R_\Psi^*(\mathcal{T})=H(\mathcal{Y}\mid\mathcal{C},\mathcal{X})$, the lowest expected log loss achievable from the complete observations. This decomposition motivates two complementary design objectives: constructing a compact time series representation that preserves relevant predictive information and encoding this representation for direct answer prediction by the LLM.

\subsection{Learning to Compress for Sufficiency}
\label{sec:compression}
The sufficiency divergence in Section~\ref{sec:reduction} quantifies the predictive information lost in dimension reduction. To reduce this divergence, we need to choose a compact representation of the time series. Theoretically, nonlinear SDR permits this representation to be any nonlinear function of the observations~\cite{lee2013general}, but the space of such functions is too large to search with limited supervision. Prior work on time series analysis provides a finite set of interpretable features that characterize time series patterns~\cite{fulcher2013highly}. 
However, most features in the full set are irrelevant to a given question, and including them as input lowers the performance of System 1 models~\cite{typesafe2026api}. TimesJev therefore compresses each series into a compact representation by selecting a small set of these features, with a surrogate objective for the sufficiency divergence~\cite{barber2003information}. To optimize this selection, we relax it into a differentiable problem with stochastic gates.

\paragraph{Constrained Objective.}
Each candidate feature is a scalar function of $\mathcal{X}$ and thus a one-dimensional reduction in the sense of SDR~\cite{cook2007fisher}.
Let $F(\mathcal{X})\in\mathbb{R}^{M}$ denote the library of $M$ candidate features computed from $\mathcal{X}$, and let $F_S(\mathcal{X})$ denote the features indexed by a subset $S\subseteq\{1,\ldots,M\}$, which form the compact representation $\mathcal{T}:=F_S(\mathcal{X})$. Formally, we seek a subset of at most $K$ features that best preserves the predictive information about the correct answer $\mathcal{Y}$, i.e., $\min_{S\subseteq\{1,\ldots,M\},\,|S|\le K}\Delta_{\mathrm{suff}}(S)$.

Since $p(\mathcal{Y}\mid \mathcal{C},\mathcal{X})$ is unknown, $\Delta_{\mathrm{suff}}$ cannot be computed directly. We instead use the cross-entropy $\mathcal L(S,\eta)=\mathbb E[-\log r_\eta(\mathcal{Y}\mid \mathcal{C},\mathcal{T})]$ of a learned predictor $r_\eta$. For any $r_\eta$:
\begin{equation}
\begin{aligned}
\mathcal L(S,\eta)={}&H(\mathcal{Y}\mid \mathcal{C},\mathcal{X})+\Delta_{\mathrm{suff}}(S)\\
&+\mathop{\mathbb{E}}\limits_{\mathcal{C},\mathcal{T}}\mathrm{KL}\bigl(p\,\Vert\,r_\eta\bigr),
\end{aligned}
\label{eq:decomp}
\end{equation}
where $p$ and $r_\eta$ in the KL term are both conditioned on $(\mathcal{C},\mathcal{T})$. The KL term is nonnegative, so $\mathcal L(S,\eta)-H(\mathcal{Y}\mid \mathcal{C},\mathcal{X})\ge\Delta_{\mathrm{suff}}(S)$ for every $r_\eta$, and the difference between the two sides is exactly the KL term, which vanishes when $r_\eta$ equals the true conditional $p(\cdot\mid \mathcal{C},\mathcal{T})$. Minimizing the cross-entropy therefore minimizes a variational upper bound on the sufficiency divergence~\cite{barber2003information}, since $H(\mathcal{Y}\mid \mathcal{C},\mathcal{X})$ is a constant independent of $S$ and $\eta$. This gives the surrogate objective $\min_{|S|\le K}\min_{\eta}\mathcal L(S,\eta)$. Training $r_\eta$ is standard, but choosing $S$ is a discrete search over exponentially many subsets. The next paragraph relaxes this search into a differentiable problem using stochastic gates.

\paragraph{Learning with Stochastic Gates.}
To optimize the discrete subset $S$, we follow~\citet{yamada2020feature} and relax the binary inclusion indicator of each feature into a continuous random variable, called a stochastic gate (STG), which is trained jointly with the variational predictor $r_\eta$.
For the $j$-th feature, the stochastic gate is:
\begin{equation}
    z_j = \operatorname{clip}(\mu_j+\epsilon_j,0,1),\quad
    \epsilon_j \sim \mathcal{N}(0,\sigma^2),
    \label{eq:stochastic_gate}
\end{equation}
where $\mu_j$ is learnable and $\sigma$ is fixed. The gates $z=(z_1,\ldots,z_M)$ are shared across examples, and the predictor receives the gated input $\widetilde{\mathcal{T}}=z\odot F(\mathcal{X})$. We then jointly optimize the gate parameters $\mu$ and the predictor parameters $\eta$ by minimizing the following objective:
\begin{equation}
    \scalebox{0.9}{$\displaystyle
    \mathcal{L}_{\mathrm{sel}}
    =\mathbb{E}\bigl[-\log r_\eta(\mathcal{Y}\mid \mathcal{C},\widetilde{\mathcal{T}})\bigr]
    +\lambda\sum_{j=1}^{M}\Phi(\mu_j/\sigma)$},
    \label{eq:stg_objective}
\end{equation}
where the expectation is over the data and gate noise, and $\Phi$ is the standard normal CDF. The first term is the cross-entropy $\mathcal L(S,\eta)$ evaluated on the gated input. The second term relaxes the constraint $|S|\le K$ into a penalty on the expected number of retained features, where $\Pr(z_j>0)=\Phi(\mu_j/\sigma)$ is the probability that the $j$-th feature is retained. The hyperparameter $\lambda$ controls its strength.
After training, we keep the $K$ features with the largest $\mu_j$, i.e., $S=\operatorname{TopK}(\mu)$ with $\mu=(\mu_1,\ldots,\mu_M)$, set $\mathcal{T}=F_S(\mathcal{X})$, and pass $\mathcal{T}$ to the encoding stage.

\subsection{Learning to Encode for Realizability}
\label{sec:embedding}
The representation selected in Section~\ref{sec:compression} lowers the sufficiency divergence, but the realizability divergence in Section~\ref{sec:reduction} also affects performance. This divergence remains because the selected temporal features are numerical values that differ in scale and meaning, while System 1 models read these values as text tokens and are less reliable on numerical input~\cite{typesafe2026api}. TimesJev therefore lowers this divergence by learning an encoder that maps the compact representation into the input embedding space, as prior work on time series embedding for LLMs~\cite{jin2024time}.

\paragraph{Feature Space Mapping.}
The selected features differ in both numerical scale and statistical meaning. Prior work uses feature-specific numerical embeddings to encode both values and feature identities~\cite{gorishniy2022embeddings}, and we build on this idea to implement such mappings through a shared encoder conditioned on feature identity.
For the $k$-th selected feature, $k=1,\ldots,K$, let $v_k$ denote its value; since the selected subset $S$ is fixed, the position $k$ identifies the feature. We transform $v_k$ into a numerical representation $\rho(v_k)\in\mathbb{R}^{4}$ containing its signed logarithmic magnitude, sign, scaled log magnitude, and bounded value. The feature identity is represented by a one-hot vector $e_{k}\in\{0,1\}^{K}$. A shared encoder $g_\theta$ maps their concatenation into the backbone's embedding space of dimension $D$:
\begin{equation}
    h_k = g_\theta\!\left(
        [\rho(v_k),e_{k}]
    \right)\in\mathbb{R}^{D}.
    \label{eq:feature_mapping}
\end{equation}
Conditioning on the identity $e_k$ allows the encoder to produce different embeddings for the same numerical value associated with different features.

For each selected feature, we retain its name and numerical text and append a value token \texttt{[v]}, whose embedding is replaced by $h_k$.
These entries, together with the question context and candidate answers, form the complete input $\psi_\theta(\mathcal{C},\mathcal{T})$.
This mapping and input construction instantiate the encoder family $\Psi$ in Definition~\ref{def:system1_sufficiency}, used to characterize the realizability divergence $\Delta_{\mathrm{real}}$. With the selected set $S$ and LLM $q_0$ fixed, we train the encoder to approach the optimal expected log loss $R_\Psi^*(\mathcal{T})$ by minimizing:
\begin{equation}
    \min_\theta\;
    \mathbb{E}\!\left[
        -\log q_0\!\left(
            \mathcal{Y}\mid\psi_\theta(\mathcal{C},\mathcal{T})
        \right)
    \right].
    \label{eq:embedding_objective}
\end{equation}
At inference, the model receives this representation to replace the raw time series, and the compact representation serves as the numerical evidence.

\section{Experiments}

\subsection{Experimental Setups}

\paragraph{Benchmarks.}
We conduct experiments on TSQA~\cite{kong2025time}, a large-scale multi-task benchmark for time series question answering, which comprises approximately 200,000 question-answer pairs across 12 domains and 5 task types. Specifically, we focus on four tasks that require a decision over a fixed set of candidate answers, matching the setting of System 1 models: anomaly detection across four domains (Environment, AIOps, Healthcare, and Web), human activity classification, multiple-choice question answering (MCQA), and true/false question answering.

\begin{table*}[ht]
\centering
\scriptsize
\setlength{\tabcolsep}{1.5pt}
\renewcommand{\arraystretch}{1.15}

\resizebox{\textwidth}{!}{
\begin{tabular}{l l cccc c cccc cccc c}
\toprule

\multicolumn{2}{c}{\multirow{2}{*}{\textbf{Method}}} &
\multicolumn{4}{c}{\textbf{Anomaly Detection}} &
\multicolumn{1}{c}{\textbf{Classification}} &
\multicolumn{4}{c}{\textbf{MCQA}} &
\multicolumn{4}{c}{\textbf{True/False QA}} &
\multirow{2}{*}{\textbf{Overall}} \\

\cmidrule(lr){3-6}\cmidrule(lr){7-7}
\cmidrule(lr){8-11}\cmidrule(lr){12-15}

\multicolumn{2}{c}{} &
Environment & AIOps & Healthcare & Web &
Human Activity &
Healthcare & Nature & Web & Other &
Healthcare & Nature & Web & Other & \\

\midrule
\multicolumn{16}{c}{\cellcolor{gray!15}\textit{General-Purpose LLMs}} \\
\multirow{2}{*}{Qwen3.5-9B} & Zero-shot
& 49.60 & 63.00 & 44.80 & 58.20 & 22.80
& 61.96 & 77.50 & 83.61 & 55.81
& 71.68 & 74.59 & 80.95 & 79.31
& 54.31 \\
\phantom{Qwen3.5-9B} & LoRA SFT
& 79.20 & 98.00 & 99.20 & 91.80 & 76.80
& \underline{71.74} & \underline{86.67} & 86.89 & 72.09
& \underline{84.27} & \underline{88.52} & 87.30 & 89.66
& 86.91 \\
\hdashline
\addlinespace[1pt]
\multirow{2}{*}{Qwen3-8B} & Zero-shot
& 50.00 & 50.40 & 54.00 & 59.80 & 21.20
& 63.04 & 66.67 & 90.16 & 53.49
& 70.98 & 80.33 & 88.89 & 72.41
& 53.91 \\
\phantom{Qwen3-8B} & LoRA SFT
& 69.20 & 97.20 & 99.00 & 85.00 & 78.40
& 68.84 & 78.33 & 86.89 & 67.44
& 77.27 & \underline{88.52} & 88.89 & 89.66
& 83.51 \\
\hdashline
\addlinespace[1pt]
\multirow{3}{*}{Time-MQA$^\dagger$} & Qwen-2.5-7B
& 52.00 & 53.20 & 97.80 & 55.20 & 9.40
& 54.35 & 68.33 & 77.05 & 55.81
& 65.38 & 75.41 & 82.54 & 89.66
& 57.09 \\
 & Mistral-7B
& 44.80 & 62.60 & 41.80 & 54.80 & 45.00
& 56.16 & 61.67 & 55.74 & 51.16
& 68.88 & 78.69 & 85.71 & 82.76
& 54.31 \\
 & Llama-3-8B
& 58.80 & 84.00 & 64.00 & 54.80 & 38.20
& 56.88 & 64.17 & 77.05 & 58.14
& 66.78 & 76.23 & 74.60 & 89.66
& 61.77 \\
\midrule
\multicolumn{16}{c}{\cellcolor{gray!15}\textit{Time Series Reasoning Models}} \\
\multirow{2}{*}{TimeOmni-1-7B} & Zero-shot
& 54.00 & 58.80 & 50.00 & 64.60 & 19.40
& 54.71 & 58.33 & 68.85 & 51.16
& 50.00 & 52.46 & 55.56 & 58.62
& 50.80 \\
 & LoRA SFT
& 73.60 & 97.80 & \underline{99.40} & 69.40 & 77.40
& 70.29 & 77.50 & \underline{93.44} & 74.42
& 78.67 & \underline{88.52} & 88.89 & 86.21
& 82.23 \\
\hdashline
\addlinespace[1pt]
\multirow{2}{*}{TS-Reasoner-7B} & Zero-shot
& 61.20 & 68.00 & 43.60 & 62.00 & 8.60
& 47.83 & 48.33 & 60.66 & 53.49
& 51.75 & 58.20 & 47.62 & 65.52
& 49.57 \\
\phantom{TS-Reasoner-7B} & LoRA SFT
& 64.80 & 98.40 & 98.00 & 83.00 & 70.80
& 61.23 & 70.83 & 85.25 & \underline{79.06}
& 73.43 & 85.25 & 85.71 & \underline{93.10}
& 80.17 \\
\hdashline
\addlinespace[1pt]
\multirow{2}{*}{VeriTime-4B} & Zero-shot
& 56.80 & 66.20 & 52.20 & 59.20 & 22.80
& 63.04 & 72.50 & 88.52 & 60.47
& 69.58 & 72.95 & 74.60 & 86.21
& 56.77 \\
\phantom{VeriTime-4B} & LoRA SFT
& 73.20 & 97.20 & \underline{99.40} & 57.60 & 77.40
& 66.30 & 82.50 & 85.25 & 69.77
& 80.07 & 86.07 & \underline{92.06} & 82.76
& 80.11 \\
\midrule
\multicolumn{16}{c}{\cellcolor{gray!15}\textit{Jev-Family Models}} \\
\multirow{2}{*}{Decider-4B} & Zero-shot
& 55.60 & 61.40 & 64.20 & 67.60 & 40.00
& 67.75 & 78.33 & 80.33 & 65.12
& 68.18 & 78.69 & 71.43 & 82.76
& 61.77 \\
\phantom{Decider-4B} & LoRA SFT
& 80.40 & 97.80 & \underline{99.40} & \underline{93.20} & 78.80
& 69.20 & 79.17 & 85.25 & 74.42
& 82.52 & \underline{88.52} & \underline{92.06} & 89.65
& 85.41 \\
\hdashline
\addlinespace[1pt]
\multirow{2}{*}{Decider-2B} & Zero-shot
& 49.40 & 54.40 & 56.60 & 61.00 & 37.60
& 59.42 & 66.67 & 78.69 & 62.79
& 53.50 & 58.20 & 46.03 & 75.86
& 53.97 \\
\phantom{Decider-2B} & LoRA SFT
& 77.80 & 98.20 & \underline{99.40} & 89.80 & 79.00
& 67.39 & 80.00 & 81.97 & 67.44
& 81.47 & 87.70 & 87.30 & 89.66
& 85.80 \\
\midrule
\rowcolor{blue!8}
\multicolumn{2}{l}{\textbf{TimesJev (+ Decider-4B)}}
& \textbf{86.60} & \textbf{99.20} & \textbf{99.60} & 92.40 & \underline{79.80}
& \textbf{78.47} & \textbf{89.16} & \textbf{95.08} & \textbf{81.39}
& \textbf{85.31} & \textbf{89.34} & \textbf{94.26} & \textbf{96.55}
& \textbf{89.78} \\
\multicolumn{2}{l}{\textbf{TimesJev (+ Decider-2B)}}
& \underline{84.20} & \underline{99.00} & \textbf{99.60} & \textbf{97.60} & \textbf{80.00}
& \underline{71.74} & 84.16 & 91.80 & 74.42
& 81.81 & 82.79 & 90.98 & \underline{93.10}
& \underline{87.02} \\

\bottomrule
\end{tabular}}

\caption{Main results on TSQA in accuracy (\%). LoRA SFT denotes LoRA fine-tuning on the TSQA training set. The best result is in \textbf{bold} and the second best is \underline{underlined}. $^\dagger$Official checkpoints fine-tuned on TSQA.}
\label{tab:main_results}
\end{table*}

\paragraph{Models.}
We compare TimesJev with zero-shot and LoRA fine-tuned baselines across general-purpose, time series reasoning, and Jev-family models: 1) general-purpose LLMs, including Qwen3.5-9B and Qwen3-8B~\cite{yang2025qwen3}, which answer with chain-of-thought reasoning; 2) time series reasoning models, including TimeOmni-1-7B~\cite{guan2026timeomni}, VeriTime-4B~\cite{zhou2026time}, and TS-Reasoner-7B~\cite{yu2025ts}, as well as the Time-MQA models built on Qwen-2.5-7B, Mistral-7B, and Llama-3-8B~\cite{kong2025time}, which are released by the original authors after fine-tuning on TSQA; and 3) Jev-family models, including Decider-2B and Decider-4B~\cite{marosi2026decider}, the open-source System 1 models of the Jev family~\cite{marosi2026decider}, which produce a decision and serve as the backbones of TimesJev.

\begin{table}[ht]
\centering
\scriptsize
\setlength{\tabcolsep}{3pt}
\setlength{\aboverulesep}{0pt}   %
\setlength{\belowrulesep}{0pt}
\renewcommand{\arraystretch}{1.3}

\resizebox{\columnwidth}{!}{
\begin{tabular}{l@{\hspace{6pt}}l c c c c}
\toprule
\multicolumn{2}{c}{\textbf{Method}} &
\textbf{Tokens} &
\textbf{TFLOPs} &
\textbf{Latency (ms)} &
\textbf{Overall} ($\uparrow$) \\
\midrule
\multicolumn{6}{c}{\cellcolor{gray!15}\textit{General-Purpose LLMs}} \\
\multirow{2}{*}{Qwen3.5-9B} & Zero-shot & 1535 & 35.97 & 12528 & 54.31 \\
\phantom{Qwen3.5-9B} & LoRA SFT & 7 & 6.20 & 102 & 86.91 \\
\hdashline
\multirow{2}{*}{Qwen3-8B} & Zero-shot & 577 & 14.79 & 4835 & 53.91 \\
\phantom{Qwen3-8B} & LoRA SFT & 7 & 5.22 & 103 & 83.51 \\
\midrule
\multicolumn{6}{c}{\cellcolor{gray!15}\textit{Time Series Reasoning Models}} \\
\multirow{2}{*}{TimeOmni-1-7B} & Zero-shot & 437 & 11.59 & 4088 & 50.80 \\
\phantom{TimeOmni-1-7B} & LoRA SFT & 7 & 4.79 & 92.2 & 82.23 \\
\hdashline
\multirow{2}{*}{TS-Reasoner-7B} & Zero-shot & 32 & 3.29 & 481 & 49.57 \\
\phantom{TS-Reasoner-7B} & LoRA SFT & 6 & 2.43 & 112 & 80.17 \\
\hdashline
\multirow{2}{*}{VeriTime-4B} & Zero-shot & 644 & 7.75 & 5360 & 56.77 \\
\phantom{VeriTime-Qwen3-4B} & LoRA SFT & 7 & 2.53 & 66.0 & 80.11 \\
\hdashline
\multirow{3}{*}{Time-MQA$^\dagger$} & Qwen-2.5-7B & 71 & 5.15 & 243 & 57.09 \\
 & Mistral-7B & 81 & 5.16 & 702 & 54.31 \\
 & Llama-3-8B & 101 & 5.15 & 862 & 61.77 \\
\midrule
\multicolumn{6}{c}{\cellcolor{gray!15}\textit{Jev-Family Models}} \\
\multirow{2}{*}{Decider-4B} & Zero-shot & $\approx$0 & 2.44 & 21.5 & 61.77 \\
\phantom{Decider-4B} & LoRA SFT & $\approx$0 & 2.44 & 23.0 & 85.41 \\
\hdashline
\multicolumn{2}{l}{\textbf{Ours (Decider-4B)}} & $\approx$0 & 3.95 & 34.1 & 89.78 \\
\cmidrule{1-6}
\multirow{2}{*}{Decider-2B} & Zero-shot & $\approx$0 & 1.09 & 15.8 & 53.97 \\
\phantom{Decider-2B} & LoRA SFT & $\approx$0 & 1.09 & 15.8 & 85.80 \\
\hdashline
\multicolumn{2}{l}{\textbf{Ours (Decider-2B)}} & $\approx$0 & 1.76 & 19.9 & 87.02 \\
\bottomrule
\end{tabular}}

\caption{Test-time cost and accuracy on TSQA. $\approx$0 indicates that the answer is read from option probabilities without decoding. Overall is the accuracy (\%) in Table~\ref{tab:main_results}. $^\dagger$Models fine-tuned on TSQA.}
\label{tab:efficiency}
\end{table}

\subsection{Main Results}
Table~\ref{tab:main_results} reports the accuracy of all methods on TSQA. In all three groups, zero-shot models achieve low accuracy, and LoRA fine-tuning improves their overall accuracy by 41\% to 62\% in relative terms. Among zero-shot models, the Jev-family models perform best on average, and zero-shot Decider-4B reaches the overall accuracy of the best Time-MQA model, which is fine-tuned on TSQA. TimesJev improves the overall accuracy of zero-shot Decider-4B and Decider-2B by 45.5\% and 64.5\% in relative terms, and both variants outperform all baselines in overall accuracy. Compared with the best zero-shot general-purpose LLM and time series reasoning model, which both generate reasoning steps before answering, TimesJev with Decider-4B improves the overall accuracy by 65.4\% and 58.3\%. Across tasks, the largest gain appears on MCQA, where TimesJev improves on LoRA fine-tuning of Decider-4B by 12.6\%. The gains on anomaly detection are smaller because all LoRA fine-tuned models reach high accuracy in two of the four domains, which leaves less room for improvement. Compared with the zero-shot Decider, TimesJev improves the overall accuracy on both model sizes, from 61.77 to 89.78 on Decider-4B and from 53.97 to 87.02 on Decider-2B, showing consistency across backbone sizes.

\subsection{Further Analysis}

\paragraph{Test-Time Efficiency.}
Table~\ref{tab:efficiency} compares the inference cost of all methods. Most zero-shot reasoning models generate hundreds to thousands of tokens per question, and LoRA fine-tuning shortens their outputs to a few tokens. Jev-family models instead read the answer from option probabilities. Compared with zero-shot general-purpose LLMs and time series reasoning models that generate long reasoning chains, TimesJev reduces latency by more than 99\%, a speedup of 120 to 630 times. TimesJev is also 7 to 43 times faster than the Time-MQA models and 2 to 6 times faster than the LoRA fine-tuned baselines. Both variants of TimesJev remain faster than all baselines outside the Jev family, and TimesJev with Decider-2B also requires the fewest TFLOPs among them.

\begin{figure}[t]
\centering
\includegraphics[width=0.95\columnwidth]{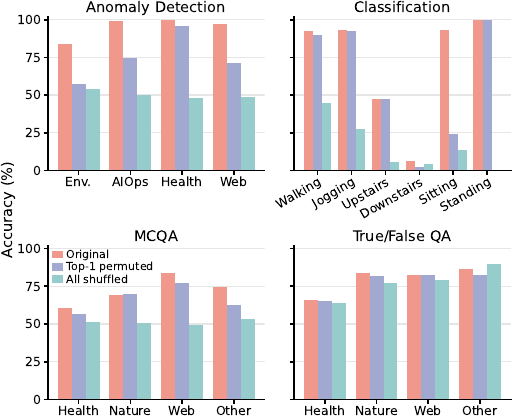}
\caption{Accuracy by domain when the feature values are kept, the most important one is permuted, or all are shuffled across questions.}
\label{fig:value_sensitivity}
\end{figure}

\paragraph{Dependence on Feature Values.}
Figure~\ref{fig:value_sensitivity} tests whether TimesJev uses the values in the compact representation. Shuffling all feature values across questions lowers the accuracy on anomaly detection and classification to about chance level, which shows that the decisions depend on the series information in the compact representation rather than on the question text. Permuting the most important feature causes large drops on three anomaly detection domains and lowers the accuracy on the sitting class by about three quarters, indicating that a few features carry most of the information for these decisions. On MCQA, shuffling all values lowers the accuracy in every domain, while permuting the top feature has a smaller effect. True/false QA changes little under both perturbations, suggesting that many of its questions can be answered from the question context.

\begin{figure}[t]
\centering
\includegraphics[width=0.95\columnwidth]{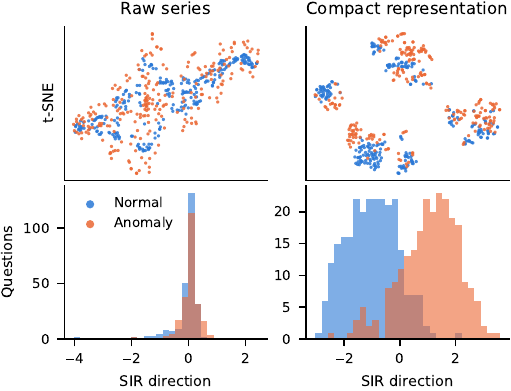}
\caption{\textbf{Top}: t-SNE embeddings of the raw series and their compact representations on Web anomaly detection. \textbf{Bottom}: sufficient summary plots along the first sliced inverse regression (SIR) direction~\citep{li1991sliced}.}
\label{fig:sufficient_summary}
\end{figure}

\paragraph{Predictive Sufficiency of Compact Representations.}
Figure~\ref{fig:sufficient_summary} visualizes the raw series and their compact representations on Web anomaly detection. In the t-SNE embeddings, the raw series form a single cloud in which normal and anomalous series are mixed, while the compact representations form several clusters. In the sufficient summary plots, we project each representation onto its first SIR direction, which is the linear projection most associated with the label. Normal and anomalous series overlap along this direction for the raw series but are largely separated for the compact representations. This result suggests that the compact representations keep the information needed to detect anomalies, consistent with predictive sufficiency. Since the SIR direction is linear, the label information in the compact representations is largely accessible through a single projection.

\begin{figure}[t]
\centering
\includegraphics[width=0.95\columnwidth]{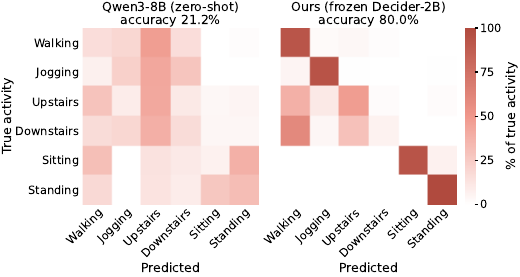}
\caption{Confusion matrices on classification for Qwen3-8B and TimesJev. Each row shows the share of a true activity assigned to each predicted activity.}
\label{fig:cls_confusion}
\end{figure}

\paragraph{Error Analysis on Classification.}
Figure~\ref{fig:cls_confusion} compares the confusion matrices of Qwen3-8B and TimesJev on human activity classification. Qwen3-8B assigns a large share of each moving activity to upstairs and spreads sitting and standing over several classes, so its accuracy stays close to chance. TimesJev reaches nearly four times its accuracy, with most predictions on the diagonal for walking, jogging and standing. The recognition of sitting relies on the median acceleration along the Z axis: permuting this feature across questions lowers the accuracy on sitting from 93.1\% to 24.1\%, while the accuracy on each other activity changes by at most 4.3\%. Its remaining errors come from the stair activities, likely because they share similar periodic movement with walking.

\section{Conclusion}
In this paper, we introduce TimesJev, a time series System 1 model. It answers each question in a single forward pass. By formalizing System 1 sufficiency and decomposing the additional prediction loss of a compressed representation into a sufficiency divergence and a realizability divergence, TimesJev enables both the selection of time series features with stochastic gates and their alignment with the input embedding space of the model. Since the System 1 model stays frozen during training, TimesJev can be added to an existing System 1 model without changing its parameters or its option readout.
Extensive experiments on multiple tasks and two System 1 backbones demonstrate improvements in overall accuracy over existing time series reasoning models with substantially lower test-time latency. Our results further highlight the importance of learning System 1 sufficient representations for efficient time series reasoning without long chains of thought.

\section*{Limitations}
While TimesJev demonstrates consistent improvements across multiple tasks and System 1 backbones, several limitations remain for future work. First, our formulation of System 1 sufficiency focuses on tasks that require a decision over a fixed set of candidate answers. Although this setting captures a common form of time series reasoning and matches the decision setting of System 1 models, extending the framework to tasks with free-form outputs such as forecasting, imputation, or open-ended question answering remains an important direction. Second, TimesJev selects time series features from a fixed library, and the stochastic gates are shared across all inputs. While the selected features are effective across the evaluated tasks and backbones, learning new features and incorporating instance-adaptive mechanisms that select features for each question may further improve flexibility. Third, TimesJev answers in a single forward pass without explicit reasoning steps, which keeps test-time cost low, and the selected features provide interpretable evidence for each decision. Generating explanations alongside these decisions for applications that require them, such as healthcare, remains an interesting direction for future investigation.

\bibliography{custom}

\appendix

\newpage
\section{TimesJev Algorithm}
Algorithm~\ref{alg:timesjev} summarizes the training and inference of TimesJev. Training consists of two stages, which reduce the two divergences in Section~\ref{sec:reduction} in turn. The first stage trains the stochastic gates jointly with the predictor $r_\eta$ and keeps the $K$ features with the largest $\mu_j$, which reduces the sufficiency divergence. The second stage fixes the selected subset $S$ and trains the encoder $g_\theta$ to map each selected feature into the input embedding space of the model, which reduces the realizability divergence. Throughout training, only the gates, the predictor and the encoder are updated, while the System 1 model stays frozen. At inference, TimesJev computes the compact representation of the test series, builds the model input, and reads the answer from the option probabilities of the model in a single forward pass.

\begin{algorithm}[t]
\small
\renewcommand{\algorithmicrequire}{\textbf{Input:}}
\renewcommand{\algorithmicensure}{\textbf{Output:}}
\caption{Training and inference of TimesJev}
\label{alg:timesjev}
\begin{algorithmic}[1]
\Require Training set $\mathcal{D}=\{(X,C,Y)\}$, feature library $F$, frozen System 1 model $q_0$, budget $K$, hyperparameters $\lambda,\sigma$
\Ensure Selected subset $S$, encoder $g_\theta$
\Statex \textit{Stage 1: Learning with stochastic gates}
\State Initialize gate parameters $\mu\in\mathbb{R}^{M}$ and predictor $r_\eta$
\Repeat
  \State Sample a mini-batch $\mathcal{B}\subset\mathcal{D}$
  \State Sample $\epsilon\sim\mathcal{N}(0,\sigma^2 I_M)$ and set $z\gets\operatorname{clip}(\mu+\epsilon,0,1)$
  \State $\widetilde{T}\gets z\odot F(X)$ for each $(X,C,Y)\in\mathcal{B}$
  \State Update $\mu$ and $\eta$ by minimizing Eq.~\eqref{eq:stg_objective} on $\mathcal{B}$
\Until{convergence}
\State $S\gets\operatorname{TopK}(\mu)$
\Statex \textit{Stage 2: Aligning the compact representation with the model}
\State Initialize encoder $g_\theta$
\Repeat
  \State Sample a mini-batch $\mathcal{B}\subset\mathcal{D}$
  \For{each $(X,C,Y)\in\mathcal{B}$}
    \State $T\gets F_S(X)$ with values $\{v_k\}$ and identities $\{\tau_k\}$
    \State $h_k\gets g_\theta([\rho(v_k),e_{\tau_k}])$ for $k=1,\ldots,K$
    \State Build $\psi_\theta(C,T)$, replacing each \texttt{[v]} embedding with $h_k$
  \EndFor
  \State Update $\theta$ by minimizing Eq.~\eqref{eq:embedding_objective} on $\mathcal{B}$ with $q_0$ frozen
\Until{convergence}
\Statex \textit{Inference}
\State Given a test question $(X,C)$, build $\psi_\theta(C,F_S(X))$
\State $\hat{Y}\gets\arg\max_{y}\,q_0\bigl(y\mid\psi_\theta(C,F_S(X))\bigr)$ in a single forward pass
\end{algorithmic}
\end{algorithm}

\section{Extended Results}

\subsection{Training Details}
Table~\ref{tab:training_data} summarizes the datasets used in our experiments, all of which come from TSQA~\citep{kong2025time}. Anomaly detection covers four domains with two options, human activity classification has six classes, and MCQA and true/false QA mix questions from five domains. The feature library contains 69 features for all datasets except HAR, which uses 240. We train TimesJev on a single NVIDIA RTX 5090 GPU.

Table~\ref{tab:training_settings} lists the training settings, which are shared by all datasets. The alignment stage trains only the value-token MLP with 0.54M parameters. We keep $K=32$ features, fix the gate noise scale at $\sigma=0.5$, and select $\lambda$ from $\{0.003, 0.01, 0.03, 0.1\}$ on held-out training data.
Training uses AdamW for three epochs with a learning rate of $10^{-3}$, and TimesJev is trained once with seed 42.

\begin{table}[t]
\centering
\resizebox{\columnwidth}{!}{%
\begin{tabular}{llcccccc}
\toprule
Task & Dataset & Opt. & Len. & $M$ & Train & Test & Time \\
\midrule
Anomaly Detection & 4 domains & 2 & 8--64 & 69 & 13,787 & 7,000 & 87.8 \\
\midrule
Classification & HAR & 6 & 30 & 240 & 4,000 & 2,000 & 28.7 \\
\midrule
MCQA & Mixed & 2--4 & 4--48 & 69 & 4,000 & 2,000 & 13.1 \\
\midrule
True/False QA & Mixed & 2 & 4--48 & 69 & 3,703 & 2,000 & 10.5 \\
\bottomrule
\end{tabular}}
\caption{Datasets and training statistics. All datasets come from TSQA; Mixed covers its five domains. Anomaly detection trains one model per domain, so its counts and time are summed over the four domains. Opt.\ is the number of options. Len.\ is the number of values in a question's series. $M$ is the size of the feature library. Time is the training time in minutes.}
\label{tab:training_data}
\end{table}

\begin{table}[t]
\centering
\small
\begin{tabular}{ll}
\toprule
Setting & Value \\
\midrule
Backbone & Decider-2B, frozen \\
Trainable module & Value-token MLP (0.54M) \\
Features & $K = 32$, stochastic gates \\
Gates & $\sigma = 0.5$, $\lambda \in \{0.003, 0.01, 0.03, 0.1\}$ \\
Optimizer & AdamW, lr $10^{-3}$, weight decay 0 \\
Schedule & 5\% warm-up, cosine decay, 3 epochs \\
Batch size & Up to 16,384 tokens \\
Seed & 42 \\
\bottomrule
\end{tabular}
\caption{Training settings, shared by all datasets.}
\label{tab:training_settings}
\end{table}

\begin{figure}[t]
\centering
\includegraphics[width=0.95\columnwidth]{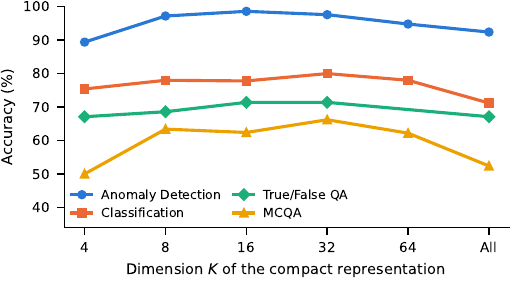}
\caption{Accuracy of TimesJev with the Decider-2B versus the number of selected features $K$.}
\label{fig:num_reductions}
\end{figure}

\paragraph{Sufficiency–Realizability Trade-off.} Figure~\ref{fig:num_reductions} shows the accuracy of TimesJev as the dimension $K$ of the compact representation increases. On all tasks, accuracy rises with $K$, peaks between 16 and 32, and falls as more features are added. A small $K$ loses information needed to answer the question, and MCQA is affected most, with accuracy at the smallest $K$ about one quarter below its peak. A large $K$ adds content unrelated to the question, and using the full library lowers MCQA by about one fifth and classification by about one tenth relative to their peaks. This trend is consistent with the two divergences in Section~\ref{sec:reduction}, since a larger $K$ reduces the sufficiency divergence but increases the realizability divergence.

\begin{figure}[t]
\centering
\includegraphics[width=0.95\columnwidth]{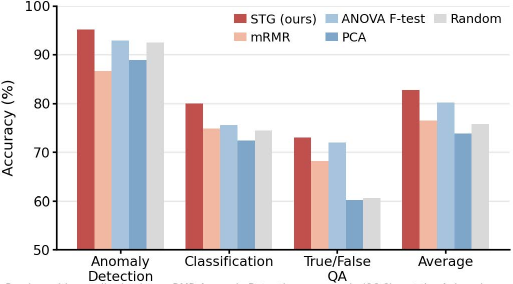}
\caption{Comparison of stochastic gates with other dimension reduction methods.}
\label{fig:reduction_methods}
\end{figure}

\paragraph{Effectiveness of Stochastic Gates.}
Figure~\ref{fig:reduction_methods} compares stochastic gates with other dimension reduction methods. STG achieves the highest accuracy on all three tasks and on average. The gains are largest on true/false QA, where STG improves over PCA and random selection by about 21\% and 20\% in relative terms. Filter methods such as mRMR and the ANOVA F-test score each feature by its statistical dependence on the label, independently of the model. In contrast, STG is trained jointly with the predictor to minimize the answer prediction loss, and therefore retains the features most predictive of the answer given the question. PCA performs worst on average, likely because its components mix the features and discard their identities, on which the encoder conditions. Descriptions of these methods can be found in Appendix~\ref{app:reduction_methods}.

\begin{figure}[t]
\centering
\includegraphics[width=0.95\columnwidth]{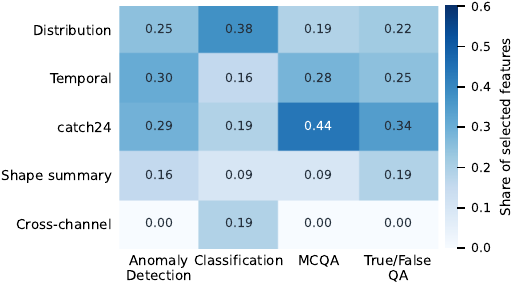}
\caption{Share of the features selected by stochastic gates that come from each family of the library.}
\label{fig:selection_heatmap}
\end{figure}

\paragraph{Composition of Compact Representations.}
Figure~\ref{fig:selection_heatmap} shows the library families of the features selected by stochastic gates. The composition differs across tasks, indicating that the gates adapt the representation to each task. Anomaly detection draws on the Temporal family most often, while classification relies most on the Distribution family and also selects cross-channel features, since its HAR series contain multiple sensor channels. MCQA and true/false QA mix questions from five domains, and catch24~\cite{lubba2019catch22} features make up more than two fifths of the selection for MCQA. Shape summaries account for the smallest share among the families shared by all tasks.

\paragraph{Ablation Study.} Table~\ref{tab:ablation} reports the ablation results with the Decider-2B. In the compression group, removing STG selection lowers accuracy by 26.7\% on MCQA and 30.5\% on true/false QA in relative terms, and removing the subset constraint lowers MCQA by 34.9\%. Replacing the compressed representation with raw observations also lowers accuracy on all four tasks, indicating that a compact representation helps the System 1 model. In the embedding group, removing value tokens causes the largest drops, ranging from 37.4\% to 56.0\% across tasks, highlighting the limitations of relying on numerical text alone. Adding the raw series to the input does not improve any task and lowers MCQA by 19.3\%.

\begin{figure}[t]
\centering
\includegraphics[width=0.95\columnwidth]{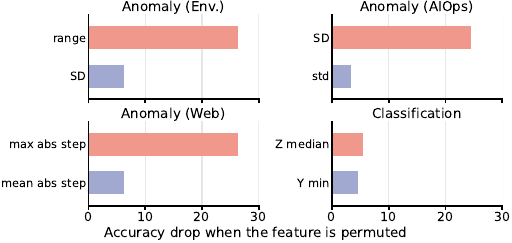}
\caption{Accuracy drop when each of the top two features is permuted across questions.}
\label{fig:permutation_importance}
\end{figure}

\paragraph{Feature Importance.}
Figure~\ref{fig:permutation_importance} shows the two features whose permutation across questions lowers the accuracy most. On each anomaly detection domain, a single feature dominates: permuting the range on Environment, the standard deviation on AIOps, or the largest absolute step on Web lowers the accuracy by 24.6 to 26.4 points, while the second feature lowers it by at most 6.2 points. On classification, the drops are smaller, at most 5.4 points for the median of the Z axis, suggesting that the decision relies on more features. Since these features have explicit meanings, the decisions on anomaly detection can be attributed to a few named properties of the series.

\begin{table*}[t]
\centering
\resizebox{0.95\textwidth}{!}{%
\begin{tabular}{l cccc c cccc cccc c}
\toprule
\multirow{2}{*}{\textbf{Variant}} & \multicolumn{4}{c}{\textbf{Anomaly Detection}} & \textbf{Classification} & \multicolumn{4}{c}{\textbf{MCQA}} & \multicolumn{4}{c}{\textbf{True/False QA}} & \multirow{2}{*}{\textbf{Overall}} \\
\cmidrule(lr){2-5}\cmidrule(lr){6-6}\cmidrule(lr){7-10}\cmidrule(lr){11-14}
 & Environment & AIOps & Healthcare & Web & Human Activity & Healthcare & Nature & Web & Other & Healthcare & Nature & Web & Other & \\
\midrule
Ours (Full) & \textbf{84.20} & \textbf{99.00} & \textbf{99.60} & \textbf{97.60} & \textbf{80.00}
& \textbf{71.74} & \textbf{84.16} & \textbf{91.80} & \textbf{74.42}   & \textbf{81.81} & \textbf{82.79} & \textbf{90.98} & \textbf{93.10}
& \textbf{87.02} \\
\midrule
\multicolumn{15}{c}{\cellcolor{gray!15}\textit{Compression}} \\
w/o STG selection & 82.00 & 98.60 & 94.60 & 87.40 & 74.40 & 57.97 & 55.83 & 73.77 & 53.49 & 61.89 & 57.38 & 61.90 & 58.62 &  70.60 \\
w/o subset constraint & 82.20 & 98.80 & 96.60 & 92.40 & 71.20 & 50.00 & 50.00 & 62.30 & 60.47 & 67.13 & 82.25 & 90.48 & 79.31 & 75.63 \\
w/o compression & 73.60 & 98.60 & 99.40 & 78.80 & 71.80 & 57.97 & 65.83 & 85.25 & 72.09 & 72.03 & 80.70 & 88.89 & 89.66 & 79.59 \\
\midrule
\multicolumn{15}{c}{\cellcolor{gray!15}\textit{Embedding}} \\
w/o value tokens & 50.20 & 51.20 & 57.20 & 47.40 & 35.20 & 48.91 & 41.67 & 31.15 & 41.86 & 51.05 & 59.84 & 57.14 & 62.07 & 48.84 \\
w/o feature identity & 84.00 & 98.20 & 92.30 & 91.60 & 78.40 & 59.06 & 66.67 & 81.97 & 58.14 & 69.93 & 81.97 & 88.89 & 90.55 & 80.13 \\
+ raw series & 80.40 & 98.00 & 94.20 & 92.40 & 78.00 & 59.42 & 70.00 & 83.61 & 60.47 & 68.88 & 81.15 & 88.89 & 82.76 & 85.51 \\
\bottomrule
\end{tabular}}
\caption{Ablation study of TimesJev with Decider-2B on TSQA in accuracy (\%). \textbf{Bold} denotes the best results.}
\label{tab:ablation}
\end{table*}

\section{Dimension Reduction Baselines}
\label{app:reduction_methods}
Figure~\ref{fig:reduction_methods} compares stochastic gates with four dimension reduction methods. Each method constructs a $K$-dimensional representation from the same library $F(X)\in\mathbb{R}^{M}$, and the rest of TimesJev, including the input encoder and the frozen model, is kept unchanged. All methods use the budget $K=32$ of the main experiments and are fitted on the same training questions as the stochastic gates.

\paragraph{mRMR.}
Minimum redundancy maximum relevance~\citep{peng2005feature} selects reductions greedily. At each step, it adds the reduction $j\notin S$ that maximizes
\begin{equation}
    I\bigl(F_j(X);Y\bigr)\Bigl(1-\frac{1}{|S|}\sum_{i\in S}\bigl|\rho\bigl(F_j(X),F_i(X)\bigr)\bigr|\Bigr),
\end{equation}
where the first factor measures the relevance to the label, the second factor penalizes the redundancy with the reductions already selected, and $\rho$ denotes the Spearman correlation.

\paragraph{ANOVA F-test.}
For each reduction, the ANOVA F-test computes the ratio of the variance between label classes to the variance within classes, and keeps the $K$ reductions with the largest F-statistics. Each reduction is scored independently of the others.

\paragraph{PCA.}
Principal component analysis~\citep{jolliffe2016principal} applies the signed logarithm $\operatorname{sign}(x)\log(1+|x|)$ to the library, standardizes it on the training set, and projects it onto the top $K$ principal components. Unlike the other methods, each component is a linear combination of all reductions rather than a single reduction. The encoder treats the $k$-th component as the $k$-th reduction: its score is the value, its index $k$, ordered by explained variance, is the identity, and it is named principal component $k$ in the text.

\paragraph{Random.}
Random selection draws $K$ reductions uniformly from the library and uses the same subset for all examples. We use a single draw with a fixed seed and, as for every method in this comparison, a single training run.

\section{Case Study}

\begin{figure*}[t]
\centering
\resizebox{0.80\textwidth}{!}{%
\includegraphics{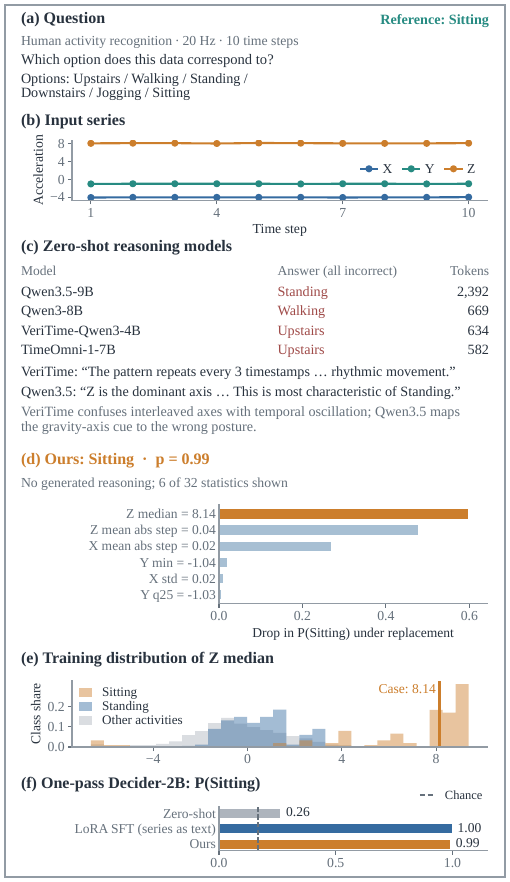}%
}
\caption{Case study on human activity classification.}
\label{fig:case_cls}
\end{figure*}

\begin{figure*}[t]
\centering
\resizebox{0.80\textwidth}{!}{%
\includegraphics{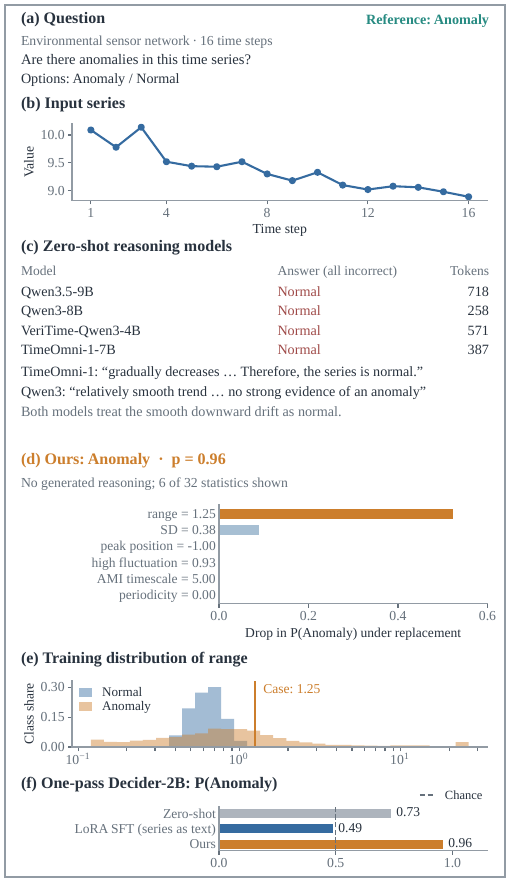}%
}
\caption{Case study on anomaly detection in the environment domain.}
\label{fig:case_env}
\end{figure*}

\begin{figure*}[t]
\centering
\resizebox{0.80\textwidth}{!}{%
\includegraphics{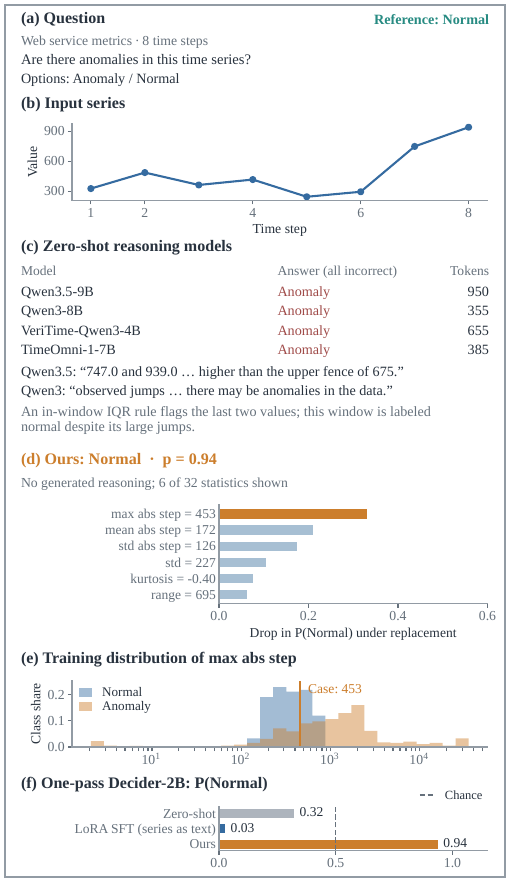}%
}
\caption{Case study on anomaly detection in the web domain.}
\label{fig:case_web}
\end{figure*}

\paragraph{Case Studies.}
Figures~\ref{fig:case_cls}, \ref{fig:case_env} and~\ref{fig:case_web} show three examples from classification and anomaly detection. In all three cases, the reasoning models generate hundreds to thousands of tokens and still answer incorrectly, as they apply generic rules to the raw values, such as reading interleaved axes as rhythmic movement, treating a smooth downward drift as normal, or flagging values above an in-window fence as anomalies. TimesJev answers all three questions correctly in a single forward pass. Replacing each selected feature shows that the decision rests on a few features, such as the median of the Z axis, the range and the largest absolute step, and panel (e) shows that their values fall in regions dominated by the correct class in the training data. In contrast, LoRA fine-tuning on the series as text fails on both anomaly detection examples.

\end{document}